%% file: root.tex
\documentclass[letterpaper, 10 pt, conference]{ieeeconf}  

\IEEEoverridecommandlockouts                              

\usepackage{graphicx}
\usepackage{amsmath}
\usepackage{dsfont}
\usepackage{subcaption}

\PassOptionsToPackage{dvipsnames}{xcolor}
\PassOptionsToPackage{table}{xcolor}

\usepackage{amssymb}
\usepackage{booktabs}
\usepackage{makecell}
\usepackage[ruled,noend,linesnumbered]{algorithm2e}
\usepackage{titlesec}
\usepackage{wrapfig}
\usepackage{array}
\usepackage{multirow}
\usepackage{colortbl}
\usepackage{tabularx}
\usepackage{macros/mysymbols}

\usepackage{grffile}
\usepackage{xspace}
\usepackage{needspace}

\usepackage{comment} 
\usepackage{url}
\usepackage{paralist}

\usepackage[belowskip=0pt,aboveskip=5pt,font=small]{caption}
\usepackage[belowskip=0pt,aboveskip=0pt,font=small]{subcaption}
\usepackage{xcolor}
\input{macros/macros.tex}

\title{\LARGE \bf
Learning Navigation Skills for Legged Robots with \\ Learned Robot Embeddings
}

\author{Joanne Truong$^{1, *}$, Denis Yarats$^{2}$, Tianyu Li$^{2}$, \\ Franziska Meier$^{2}$, Sonia Chernova$^{1, 2}$, Dhruv Batra$^{1, 2}$, Akshara Rai$^{2}$
\thanks{$^*$Work done while at FAIR}
\thanks{$^{1}$Georgia Institute of Technology, {\tt\small \{truong.j, dbatra, chernova\}@gatech.edu}}%
\thanks{$^{2}$Facebook AI Research {\tt\small \{akshararai, fmeier, tianyul, denisy\}@fb.com}}%
}

\begin{document}
\bstctlcite{IEEEexample:BSTcontrol}

\maketitle
\thispagestyle{empty}
\pagestyle{empty}

\begin{abstract}
Recent work has shown results on learning navigation policies for idealized cylinder agents in simulation and transferring them to real wheeled robots. 
Deploying such navigation policies on legged robots can be challenging due to their complex dynamics, and the large dynamical difference between cylinder agents and legged systems. In this work, we learn hierarchical navigation policies that account for the low-level dynamics of legged robots, such as maximum speed, slipping, contacts, and learn to successfully navigate cluttered indoor environments. 
To enable transfer of policies learned in simulation to new legged robots and hardware, we learn dynamics-aware navigation policies across multiple robots with robot-specific embeddings. The learned embedding is optimized on new robots, while the rest of the policy is kept fixed, allowing for quick adaptation. We train our policies across three legged robots in simulation - 2 quadrupeds (A1, AlienGo) and a hexapod (Daisy). At test time, we study the performance of our learned policy on two new legged robots in simulation (Laikago, 4-legged Daisy), and one real-world quadrupedal robot (A1). Our experiments show that our learned policy can sample-efficiently generalize to previously unseen robots, and enable sim-to-real transfer of navigation policies for legged robots. 
\end{abstract}


\input{intro}
\input{background}

\input{method}

\input{experiments}

\input{conclusion}

\input{acknowledgements}

\bibliography{references}{}
\bibliographystyle{IEEEtran}

\end{document}

%% file: macros/macros.tex
\newcommand{\xhdr}[1]{\vspace{2pt}\noindent\textbf{#1}}

\newcommand{\tableTitle}[1]{\normalsize{#1}}%
\newlength{\figwidth}%



%% file: intro.tex
\section{Introduction}
\begin{figure}[t]
  \centering%
  \resizebox{\columnwidth}{!}{
  \renewcommand{\tableTitle}[1]{\huge{#1}}%
  \setlength{\figwidth}{0.3\columnwidth}%
  \setlength{\tabcolsep}{1.5pt}%
  \renewcommand{\arraystretch}{0.8}%
  \renewcommand\cellset{\renewcommand\arraystretch{0.8}%
  \setlength\extrarowheight{0pt}}%
  
    \begin{tabular}{c c}
   \makecell{\includegraphics[width=0.25\textwidth]{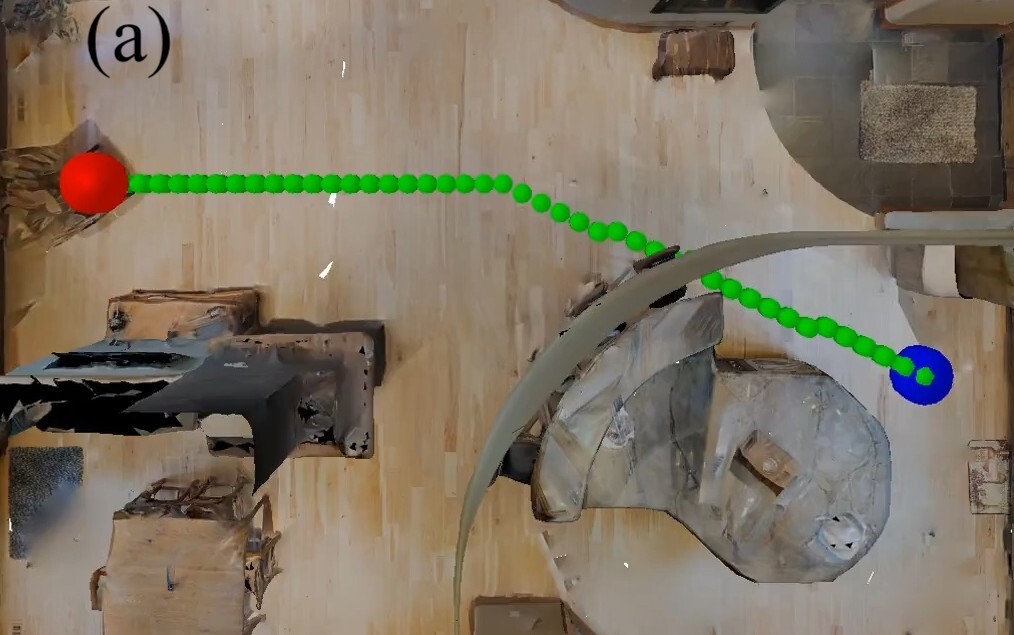}} &%
   \makecell{\includegraphics[width=0.25\textwidth]{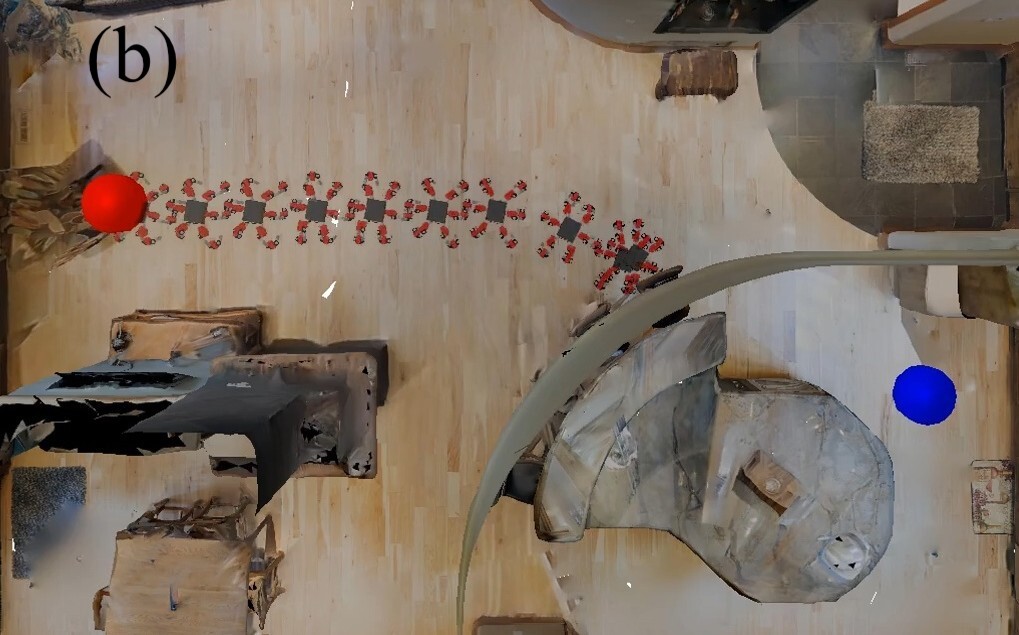}} \\
   \makecell{\includegraphics[width=0.25\textwidth]{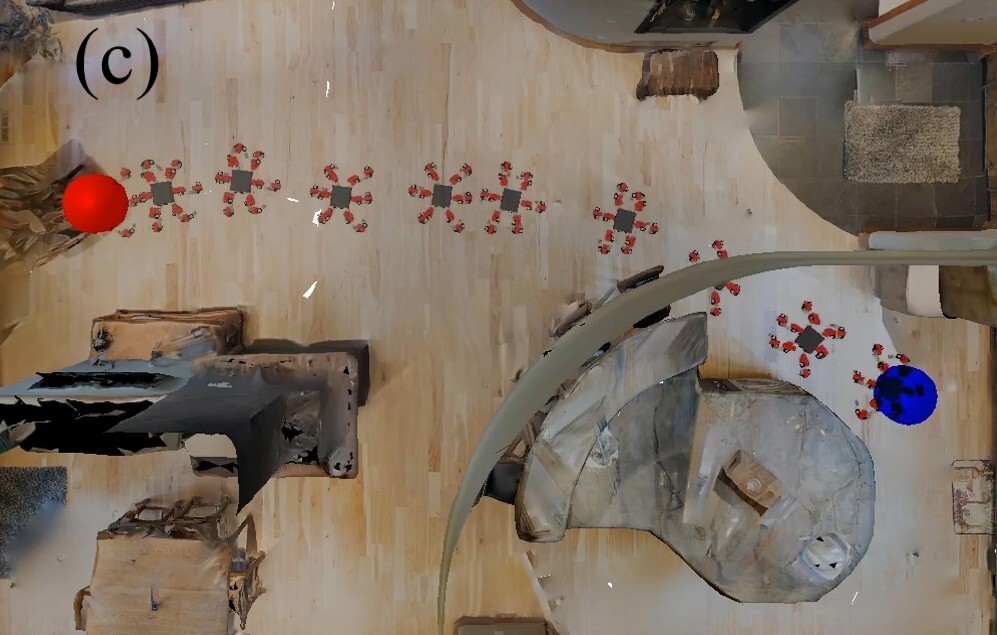}} &
   \makecell{\includegraphics[width=0.25\textwidth]{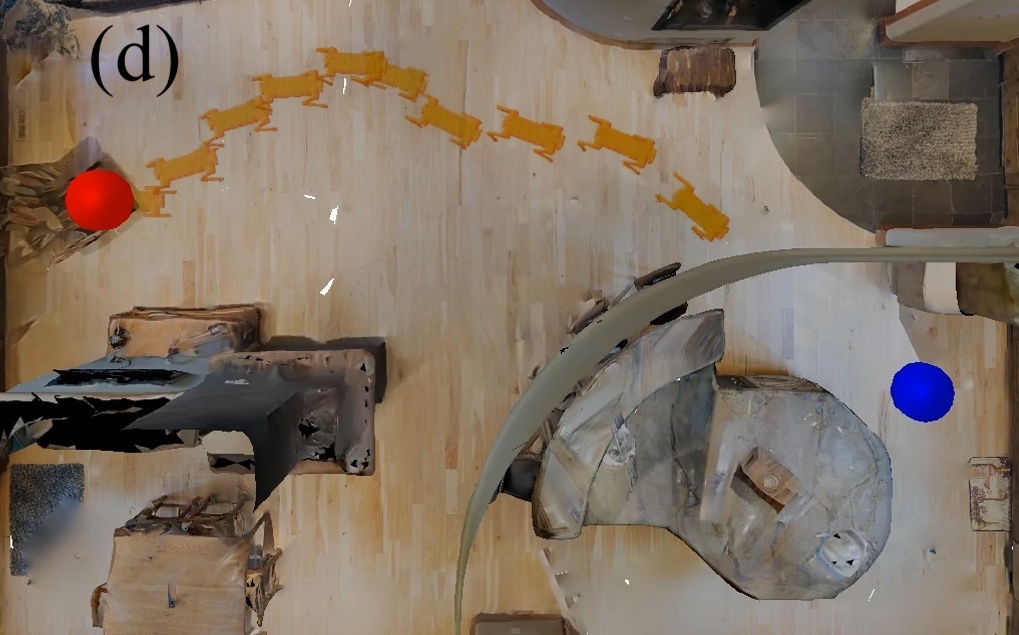}} \\
   \multicolumn{2}{c}{\includegraphics[width=0.51\textwidth]{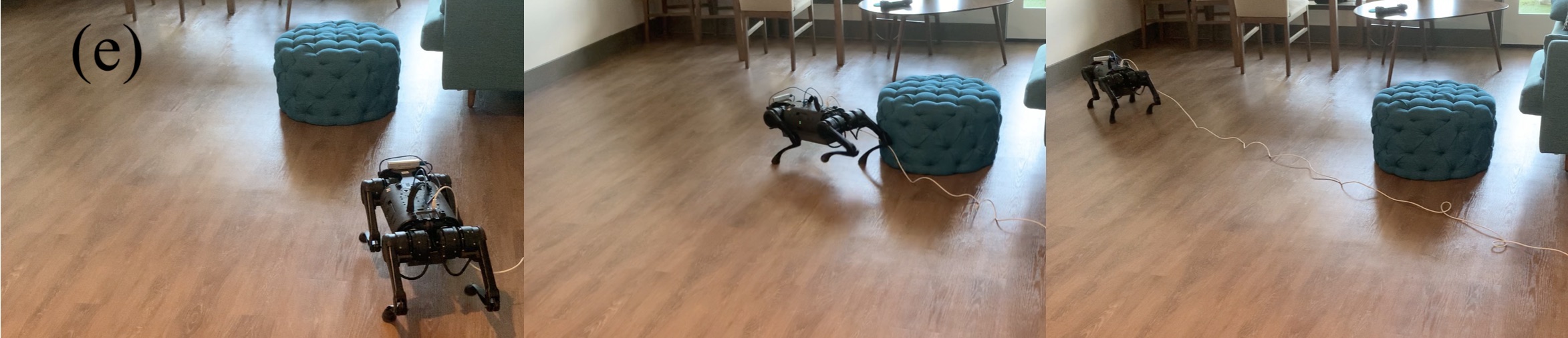}}
  \end{tabular}}
  \caption{\small Challenges of indoor navigation with legged robots. While an idealized spherical agent (a) can navigate to the goal with a PointNav policy, the same policy on Daisy (b) results in the robot getting stuck around an obstacle. A navigation policy trained on Daisy, with just egocentric images and no maps, can navigate to the goal successfully (c), but the same policy on the AlienGo quadruped (d) results in AlienGo drifting, and not reaching the goal in maximum allowed steps. We learn navigation policies across multiple legged robots and transfer to a real A1 robot (e).
  }
  \vspace{-0.8cm}
  \label{fig:teaser}
\end{figure}

Legged robots such as quadrupeds from Boston Dynamics \cite{Spot}, Unitree \cite{Unitree} and hexapods from Hebi robotics \cite{hebi} have emerged on the market as mature, robust, commercial robotic platforms. Legged robots are agile and can easily traverse uneven ground in homes like carpets and stairs, making them ideal for indoor navigation. At the same time, progress has been made in the field of learned indoor navigation using an egocentric camera input, without the use of maps \cite{habitat19iccv, wijmans2019dd}, showing real-world generalization of learned policies \cite{bansal2020combining, kadian2020sim2real}. However, learned visual navigation literature typically considers idealized virtual cylindrical agents, and does not take the dynamics of the robot into account. While such policies can be successful on wheeled robots, legged robots have significantly different dynamics than an idealized agent, or even other legged robots. These differences arise due to properties such as turning radius, motor strength, size and mass of the robot, etc. Fig. \ref{fig:teaser}b shows an example where a PointGoal Navigation (PointNav) \cite{anderson2018evaluation} policy that assumes an idealized agent is applied on the Daisy hexapod. While the policy plans to turn around an obstacle, the robot gets stuck, due to its larger size and turning radius, which were not accounted for by the PointNav policy. On the other hand, when the policy is trained directly on Daisy (Fig. \ref{fig:teaser}c), and thus is aware of the dynamics of the robot, the policy starts turning sooner to accommodate the robot body and turning radius, leading to a successful navigation around the obstacle. Applying a policy that was learned on Daisy on another legged robot - the AlienGo quadruped (Fig. \ref{fig:teaser}d) leads to the robot drifting away, and not reaching the goal in maximum allowed steps. This example highlights the importance of incorporating robot-specific dynamics when learning navigation policies for legged robots and the challenges of generalizing such learned policies to new robots. If navigation policies learned on one legged robot do not directly generalize to new systems, the experimental cost of learning navigation policies for legged robots can be prohibitively high, especially on hardware, reducing the overall scalability of such an approach. Learning navigation policies for complex legged robots, as well robustly transferring such learned policies to hardware and new robots are important challenges in the field of autonomous navigation. 
%

\begin{figure*}[ht]
  \centering%
  \resizebox{\textwidth}{!}{
  \renewcommand{\tableTitle}[1]{\huge{#1}}%
  \setlength{\figwidth}{0.3\textwidth}%
  \setlength{\tabcolsep}{1.5pt}%
  \renewcommand{\arraystretch}{0.8}%
  \renewcommand\cellset{\renewcommand\arraystretch{0.8}%
  \setlength\extrarowheight{0pt}}%
  
  
  \hspace{-0.25cm}\begin{tabular}{c c c c c c}
  \multicolumn{3}{c}{\bfseries \tableTitle{Train}} & \multicolumn{3}{c}{\bfseries \tableTitle{Test}}\\
  \cmidrule(l{4pt}r{4pt}){1-3} \cmidrule(l{4pt}r{4pt}){4-6}\\
   \makecell{\includegraphics[height=0.19\textheight]{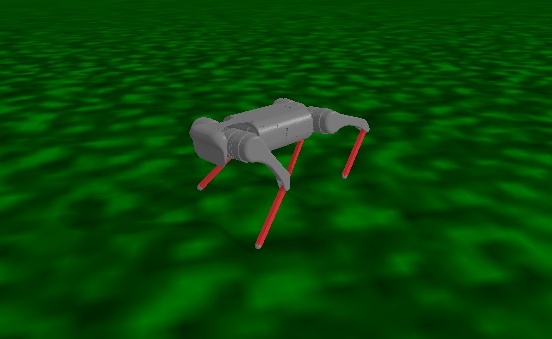}} &%
   \makecell{\includegraphics[height=0.19\textheight]{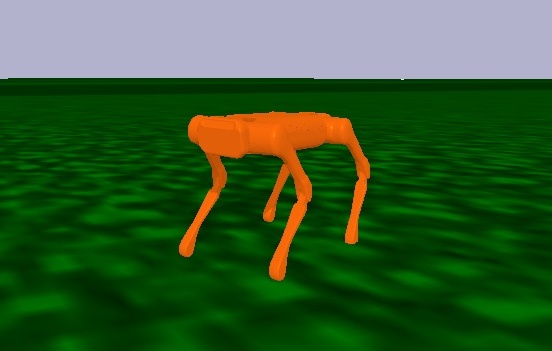}} &
   \makecell{\includegraphics[height=0.19\textheight]{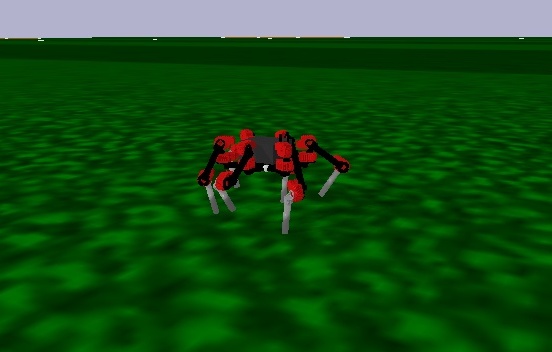}} &
   \makecell{\includegraphics[height=0.19\textheight]{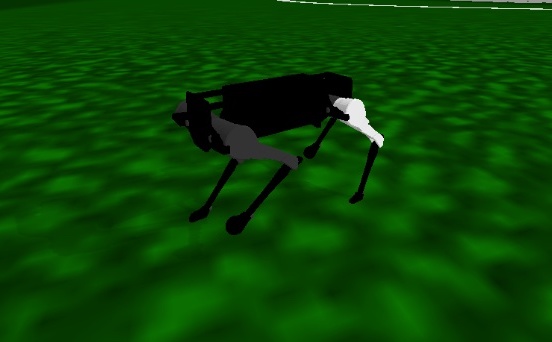}} & 
   \makecell{\includegraphics[height=0.19\textheight]{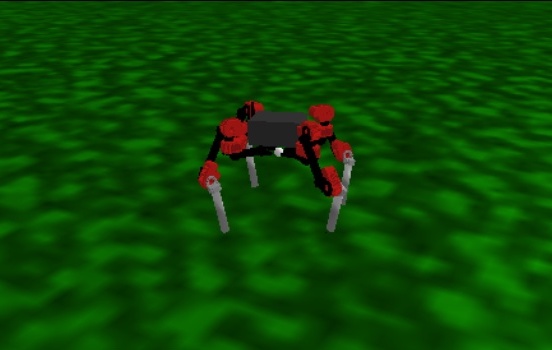}} &
   \makecell{\includegraphics[height=0.19\textheight]{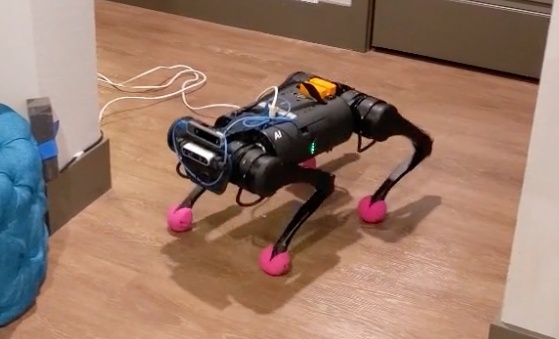}} \\
   \makecell{\includegraphics[height=0.25\textheight]{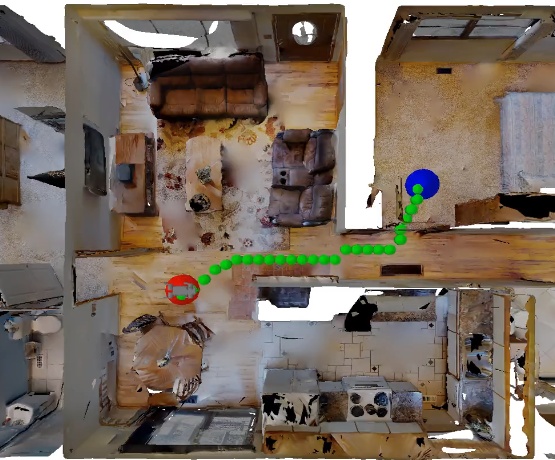}} &
   \makecell{\includegraphics[height=0.25\textheight]{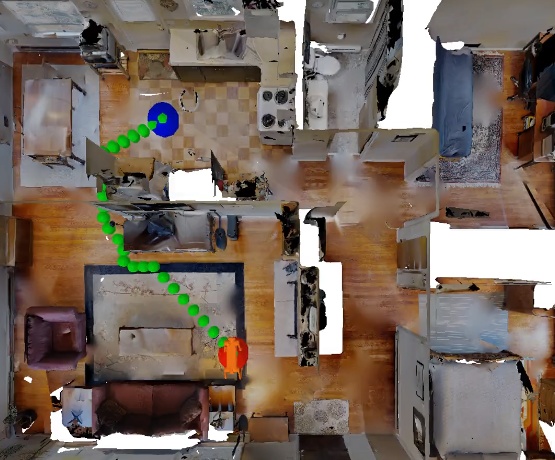}} &
   \makecell{\includegraphics[height=0.25\textheight]{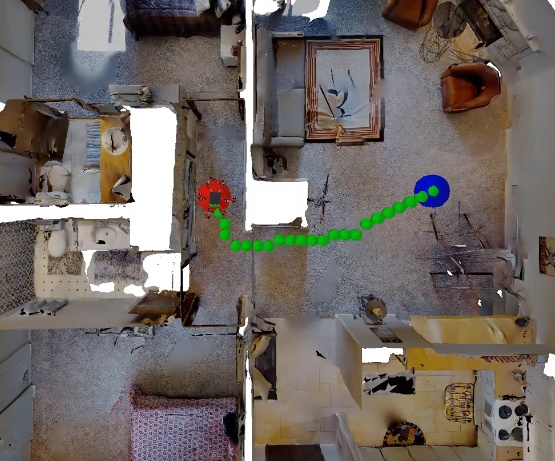}} &
   \makecell{\includegraphics[height=0.25\textheight]{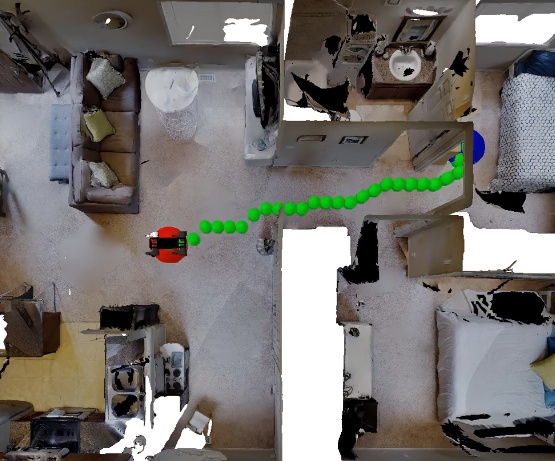}}& 
   \makecell{\includegraphics[height=0.25\textheight]{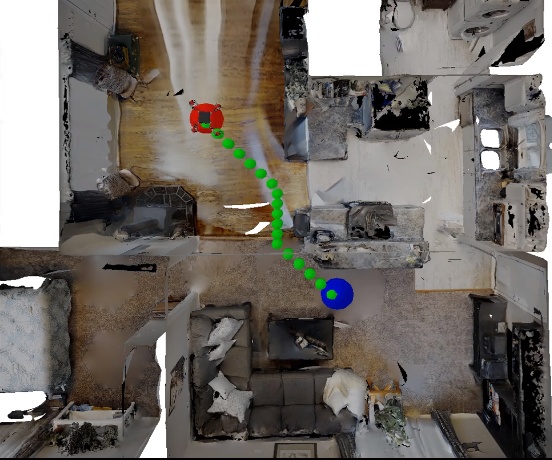}} &
   \makecell{\includegraphics[height=0.25\textheight]{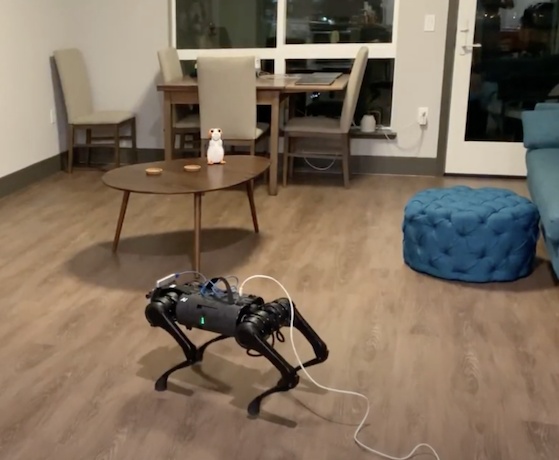}}\\
   \tableTitle{A1} & \tableTitle{AlienGo} & \tableTitle{Daisy} & \tableTitle{Laikago} & \tableTitle{4-legged Daisy} & \tableTitle{A1-Real}
  \end{tabular}}
  \caption{\small Legged robots and environments used for training and testing. We train PointNav policies in cluttered environments on A1, AlienGo, and Daisy. At test-time, we study generalization to Laikago and 4-legged Daisy in simulation, and A1 in the real-world.
  }
  \label{fig:robots}
  \vspace{-0.7cm}
\end{figure*}
In this work, we take a step towards learning visual navigation policies for legged robots, which use egocentric camera inputs and no maps, and generalize to hardware and new legged systems. We learn a navigation policy across three legged robots -- two quadrupeds (A1, AlienGo) and one hexapod (Daisy), in simulated indoor home environments (Fig. \ref{fig:robots}), and test on previously unseen simulated legged robots (Laikago, 4-legged Daisy) and a real quadruped robot (A1-Real) in new environments. We develop hierarchical navigation policies that divide the control of the robot into a high-level policy that reasons about the center of mass motion and a low-level policy that converts high-level (velocity) commands into desired footsteps. 
The high-level policy consists of two components: (1) a universal navigation policy (shared across multiple robots) and (2) a learned robot-specific embedding used to specialize the universal policy per robot. 
During training, we collect data from multiple robots in separate environments which is collectively used to update a universal navigation policy, thereby improving sample-efficiency and leading to a policy that is inherently robust to variations in robot dynamics. 
Once trained, this high-level policy can be used on new legged robots in unseen environments, by searching in the learned embedding space, keeping the universal navigation policy fixed. To the best of our knowledge, this is the first work that (1) demonstrates learning navigation policies purely from visual input across multiple legged robots; and (2) generalizes learned policies to hardware and previously unseen simulated robots. 
\newline
\indent The core contribution of this work is a sample-efficient approach for teaching legged robots to navigate unknown indoor environments. 
Our simulation results are presented in the iGibson~\cite{xia2020interactive} photo-realistic simulation, with a Pybullet~\cite{pybullet} physics engine. We also present zero-shot sim-to-real generalization hardware experiments on an A1 robot (Fig. \ref{fig:teaser}e), where the robot navigates a new home using a policy learned purely in simulation.  

%% file: background.tex
\section{Related work}
  
Navigation in unknown environments has been widely studied in robotics. We mention some closely related works.

\noindent \textbf{Indoor Navigation: }
Classical navigation approaches typically decompose the problem into mapping the environment, localizing in the map, and planning a path to the goal \cite{durrant2006simultaneous, thrun2002probabilistic, fuentes2015visual}. However, such approaches are highly dependent on the quality of the map. Mappers that only utilize RGB-D images are prone to failure in texture-less regions, and may require the use of expensive LiDAR sensors \cite{hess2016real}. Incomplete maps with errors often lead to poor performance when used by planners \cite{bansal2020combining}. Additionally, the map building process can be time-consuming, and require user-designed semantics for more robust navigation \cite{bowman2017probabilistic}. In this work, we use a learned policy that learns to extract relevant information from depth images, without explicitly building a map.

Recent works have shown learning-based approaches to be a robust at navigating unseen environments without explicit mapping (purely from egocentric RGB-D images and localization sensors) \cite{habitat19iccv, wijmans2019dd, bansal2020combining}. They can outperform traditional planning approaches \cite{bansal2020combining}, and achieve near-optimal performance in unknown houses \cite{wijmans2019dd}. However, these approaches have only been demonstrated on simple, cylindrical mobile base robots \cite{kadian2020sim2real}, and drones \cite{sadeghi2016cad2rl}. In contrast to these robots, legged robots have significantly more complex dynamics, arising from contacts and interactions between the robot and the environment, and policies learned in simulation do not directly generalize to hardware, due to the sim-to-real gap. In this work, we learn hierarchical navigation policies for legged robots that generalize to hardware and new robots. 

\noindent \textbf{RL for planning in legged robots: }
There is a significant body of work on teaching legged robots to navigate rough terrain, and robustly respond to disturbances in the environment \cite{martin2017experimental, feng, Lee_2020, li2020planning, hubicki2018walking}. Some of these works build policies that can navigate uneven terrain without any visual input \cite{martin2017experimental, hubicki2018walking, Lee_2020}, while others use LiDAR sensors or depth cameras to map the terrain, and use it to choose footstep locations \cite{feng, kim2020vision, tsounis2020deepgait}. However, these works do not address challenges associated with indoor navigation, where the robot has to navigate around obstacles, go through narrow hallways, and reach far away goals, without access to a map. In our experiments, the robot is placed in a new house that it has never seen before, and has to navigate to new goals using only onboard RGB-D cameras, and localization sensors, without a LiDAR, or a map. While previous work has used a LiDAR sensor for mapping the environment and using the map for navigating confined spaces \cite{buchanan2021perceptive}, we only use an onboard RGB-D camera in our experiments, making our system cheaper and more scalable.

We diverge from classical planning literature \cite{buchanan2021perceptive, fankhauser2018probabilistic, kim2020vision} by not building maps. Instead, we use a reactive policy trained using RL to reason about the space around the robot, avoid obstacles as it observes them, and navigate to a goal. The policy is trained in photo-realistic simulations of scans of different homes, and can be applied to a real robot operating in a new home. This helps with challenges such as building expensive maps \cite{hornung2013octomap}, planning errors from inaccurate maps \cite{al2020accuracy}, and inflating obstacles for trajectory optimization \cite{grey2017footstep}.    


\noindent \textbf{Context-aware transfer learning: }
To adapt to new settings such as sim-to-real transfer, several works propose a context-aware learning setup, where a latent representation of context is learned over a large range of training tasks. At test time, the learned latent representation is used to infer the new context~\cite{duan2016rl, rakelly2019efficient}, or optimized for generalization~\cite{peng2020learning, yu2019sim, yu2017preparing}. \cite{peng2020learning, yu2019sim} identify and encode dynamics parameters of a robot into a latent representation and learn a policy conditioned on this latent input. 
However, these approaches require expert knowledge to determine which dynamics parameters to encode, which can be difficult to choose when learning policies across robots of different morphologies. 
Instead, we propose to fully learn an embedding space for each robot, obviating the need for expert domain knowledge.

%% file: method.tex
\section{Dynamics-aware navigation policies}
\begin{figure}[t]
    \centering
    \includegraphics[width=0.5\textwidth]{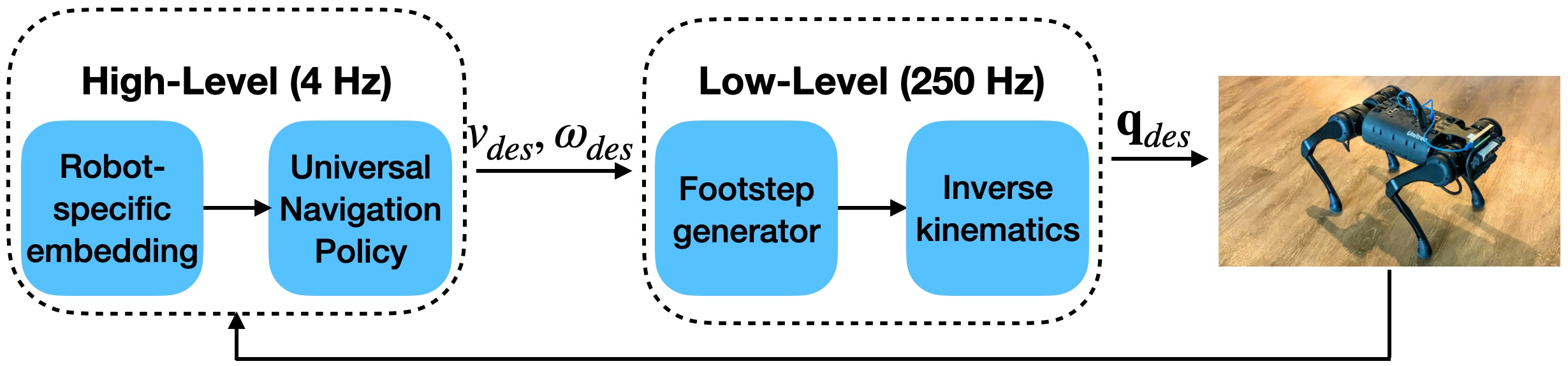}
    \caption{\small A flowchart describing our hierarchical controller. The high-level policy, described in Section \ref{sec:nav_policy}, commands a desired CoM linear ($v_{des}$) and angular velocity ($\omega_{des}$). The low-level controller generates a foot trajectory followed using Inverse Kinematics, which commands desired joint angles $q_{des}$ to the robot. 
    }
    \label{fig:hierarchy}
    \vspace{-0.7cm}
\end{figure}
In this section, we introduce our approach which learns a dynamics-aware navigation policy that can enable sim-to-real transfer and generalize to unseen robots and environments.

\subsection{Task: PointGoal Navigation}
We consider the task of PointGoal Navigation (PointNav), as defined by \cite{anderson2018evaluation}, in which a robot is initialized in an unseen environment and must navigate to a goal location without being provided a map. An episode is considered successful if the robot can reach within 0.36m of the goal location (approximately half the width of our largest robot) in less than 150 robot steps. 
In our experiments, the robot is equipped with an egocentric depth sensor and an egomotion estimator for localization. The robot has access to the goal coordinates relative to the robot, specified as polar coodinates $(r, \theta)$, where $r$ is the Euclidian distance to the goal, and $\theta$ is the relative angle to the goal. The environments consist of cluttered indoor spaces and the robot has to sense and reason about moving around obstacles to reach a randomly selected goal up to 7m away. For evaluation, we consider the success rate of our policy at reaching random new goals, as well as, Success inversely weighted by Path Length (SPL) \cite{anderson2018evaluation} to measure the efficiency of the path taken. 

\subsection{Dynamics-aware navigation policy}
\label{sec:nav_policy}

Our dynamics-aware navigation policy consists of a two-layer hierarchical controller that divides the control of legged robots into a high-level policy that commands linear and angular center of mass (CoM) velocities, and a low-level policy that achieves these desired commands (Fig. \ref{fig:hierarchy}). 

\xhdr{High-level policy:}
The high-level policy consists of (1) a universal navigation policy shared across robots, and (2) a robot-specific embedding. The universal navigation policy takes as input the observed state $s$ consisting of an egocentric depth image, the goal coordinates (in robot's reference frame), and a robot-specific embedding $z$ (Section \ref{sec:up}). The output of the policy, or action $a$ is a 2-dimensional vector containing the desired CoM linear (forward or backward) and angular (yaw) velocities  $(v_{des}, \omega_{des})$ for the robot to follow. Note that though our policy operates in a horizontal plane with 3 degrees of freedom (x-y and yaw), it only commands a 2-dimensional velocity vector, making the problem underactuated and non-holonomic. To move laterally, the robot first needs to turn and then command a linear velocity along its new heading. This makes learning the high-level policy more challenging, but once learned, the policy is easier to generalize across multiple robot morphologies.

We use soft-actor critic (SAC) with an Autoencoder (AE) (SAC+AE) \cite{yarats2019improving} as the off-policy training approach of our choice (training details described in Section \ref{sec:training}). However, unlike \cite{yarats2019improving}, we use egocentric depth images from the robot's camera, instead of a physically implausible floating 3rd-person camera. Since the camera is mounted on the robot, the image input changes as the robot moves through space, for example as the robot base tilts, the policy has to be robust to such dynamic camera movements. 

\xhdr{Low-level policy:}
\label{sec:low_level}
Our low-level policy converts the desired CoM velocities $(v_{des}, \omega_{des})$ commanded by the high-level policy into desired footstep locations using an expert designed feedback policy, similar to \cite{raibert}, followed using inverse kinematics (IK). The stance legs maintain a fixed CoM height and achieve CoM velocity $(v_{des}, \omega_{des})$ using joint position control in simulation, and whole-body MPC \cite{kim2019highly} on hardware. 

%
%

One physical robot step takes $\Delta t=0.25$s; during this time, the robot's foot moves from current location $(x_{cur}, y_{cur})$ to a desired foot placement location $(x_{des}, y_{des})$. Given $(v_{des}, \omega_{des})$ from the high-level policy, we calculate the desired change in footstep position $(\delta x_{f}, \delta y_{f})$ using the desired change in CoM position $\delta x_{com} = v_{des} \Delta t$ and orientation $\delta \gamma_{com} = \omega_{des} \Delta t$. For turning, the foot moves in a circle of radius $r_f$, changing the current angle between the foot and the robot heading $\gamma_{cur}$ to $\gamma_{des} = \gamma_{cur} + \delta \gamma_{com}$ at the end of the step.
\begin{align}
    \delta x_{f} = \delta x_{com} + r_f \cdot (\cos(\gamma_{des}) - \cos(\gamma_{cur}))\\
    \delta y_{f} = r_f \cdot (\sin(\gamma_{des}) - \sin(\gamma_{cur})) \\
         x_{des} = x_{cur} + \delta x_{f} \quad y_{des} = y_{cur} + \delta y_{f}
\end{align}
where $r_f$ is the distance between the foot and CoM. 

\begin{figure}[t]
  \centering%
  \resizebox{\columnwidth}{!}{
  \renewcommand{\tableTitle}[1]{\large{#1}}%
  \setlength{\figwidth}{0.3\columnwidth}%
  \setlength{\tabcolsep}{1.5pt}%
  \renewcommand{\arraystretch}{0.8}%
  \renewcommand{\cellset}{\renewcommand\arraystretch{0.8}%
  \setlength\extrarowheight{0pt}}%
  
  
  \hspace{-0.25cm}\begin{tabular}{c c}
   \makecell{\includegraphics[width=0.35\textwidth]{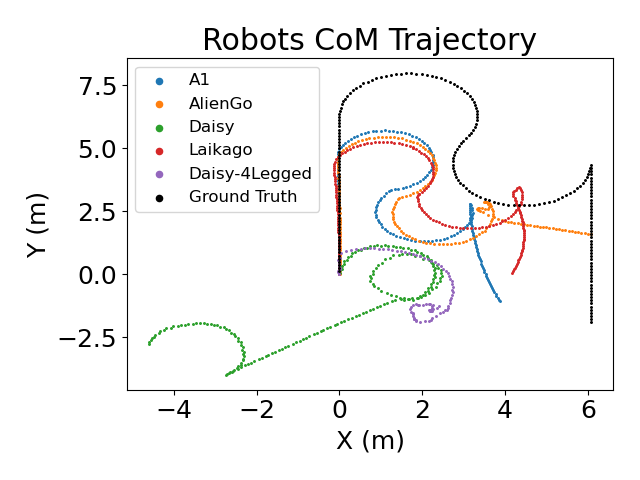}} &
   \makecell{\includegraphics[width=0.35\textwidth]{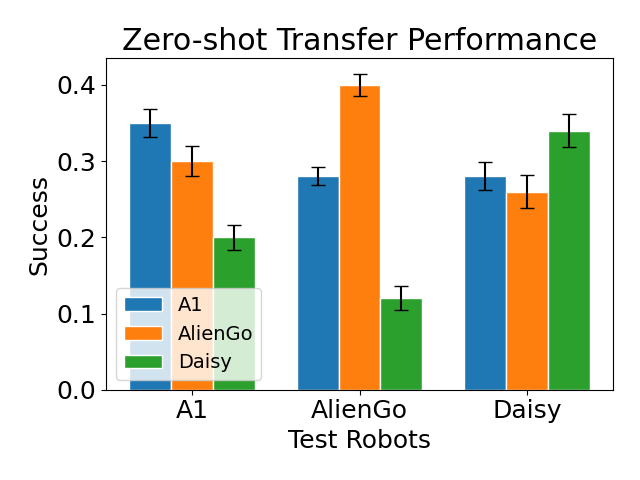}} \\
  \end{tabular}}
  \caption{\small (left): Center of Mass (CoM) trajectories for different robots following the same high-level commands. Due to the differences in the low-level controller, each robot ends up taking a different trajectory, with A1, AlienGo, Laikago very different from Daisy and 4-legged Daisy. 
  (right) Zero-shot transfer of policies trained on a single robot onto other robots. Each robot's success rate is highest with a policy trained on itself, and success rate deteriorates when using a policy trained on a different robot.
  }
  \label{fig:com_zero_shot}
  \vspace{-0.5cm}
  \end{figure}
Our footstep planner is shared across all our robots, including hexapods and quadrupeds. The footstep trajectory is then followed using robot-specific IK in swing, while stance legs maintain a fixed CoM height and desired velocity $(v_{des}, \omega_{des})$. This enables us to experiment with 5 different legged robot designs, which is otherwise cumbersome due to robot-specific controllers. As our low-level policy structure is shared across multiple robots, it is not perfect at following the high-level commands for all robots. In fact, our robots can fall when commanded high velocities, and don't follow the commands well when there is a large jump in the desired velocities. 
Fig. \ref{fig:com_zero_shot}a shows the CoM trajectories that different robots exhibit given the same high-level action sequence using our low-level controller. While A1, AlienGo and Laikago have similar trajectories in the start, they diverge later. On the other hand, Daisy and 4-legged Daisy have significantly different trajectories, and tend to turn to the left using our low-level controller. Fig. \ref{fig:com_zero_shot}b shows the success rate of reaching new goals, when pairing different low-level controllers of each robot with high-level policies trained on other robots. Each robot achieves the highest success rate when using a high-level policy that was trained with its own low-level controller, and success rate deteriorates when using high-level policies trained on other robots; the largest drop is seen in AlienGo's performance -- from $0.40$ success down to $0.26$ success when using its own policy, vs. Daisy policy, due to the largest dynamical shift in the robots. 
These results empirically confirm our hypothesis that each legged robot has different dynamics that need to be accounted for by the high-level policy in order to achieve good navigation performance, even when the low-level controllers share a common structure. In the next section, we will describe how to learn such a policy, while sharing data across robots. 

\subsection{Learning the high-level policy}
\label{sec:up}
To account for the differences between legged robots, we propose to learn a robot-specific embedding, along with a universal navigation policy across multiple robots. This allows for robot-specific specialization, while sharing data, and improved sample-efficiency.

We formulate our high-level policy $\pi_\theta$ to take as input both the observed state $s$ and a latent embedding $z_i$ per robot $i=1,2, \cdots N$, where $N$ is the total number of robots used during training. The output action (desired CoM velocities) is a function of the state and the robot-specific embedding: $a \sim \pi_\theta(a \mid s, z_i)$. The critic is formulated similarly, where the predicted Q-value is a function of the state-action pair as well as $z_i$: $Q_{\phi}(s, a, z_i) = r(s,a) + \bar{V}(s', z_i)$. $Q_{\phi}$ is the critic, $r$ is the reward, $s'$ is the next state after taking action $a$, and $\bar{V}(s', z_i)$ is the robot-specific target value function \cite{haarnoja2018soft}.

\xhdr{Robot specific embeddings: }
We formulate the robot-specific embedding as a feed-forward network $g$ with 3 layers and 100 hidden units, called the \textit{z-network}. The `input' to the z-network is a scalar, initially randomly sampled from a uniform distribution in $[0, 1)$, and learned alongside the parameters $\psi$ of the z-network. Its output is
a 1-dimensional latent embedding $z = g_\psi$, where $g$ is the z-network MLP, and $\psi$ are its learned parameters. We learn $N$ z-networks for $N$ robots, resulting in learned parameters $\psi_1, \psi_2, \cdots \psi_N$, and corresponding robot embeddings $z_1, z_2, \cdots z_N$.

\xhdr{Shared universal policy:}
We also train a shared actor $\pi_\theta$ and critic $Q_\phi$ across all robots. At the start of training, $N$ robots are initialized in different environments, and tasked with navigating to a goal location. At every step, each robot receives an egocentric depth observation, the goal vector relative to its current position, and its current robot-specific embedding $z_{1}, \ldots, z_{N}$, which is added to individual replay buffers $\mathcal{D}_i = \{(s, a, r, s', z_i)\}$; combined to create a shared replay buffer $\mathcal{D} = \{ \mathcal{D}_{1}, \mathcal{D}_{2}, $\ldots$, \mathcal{D}_{N}\}$. 

\begin{figure}[t]
    \centering
    \includegraphics[width=\columnwidth]{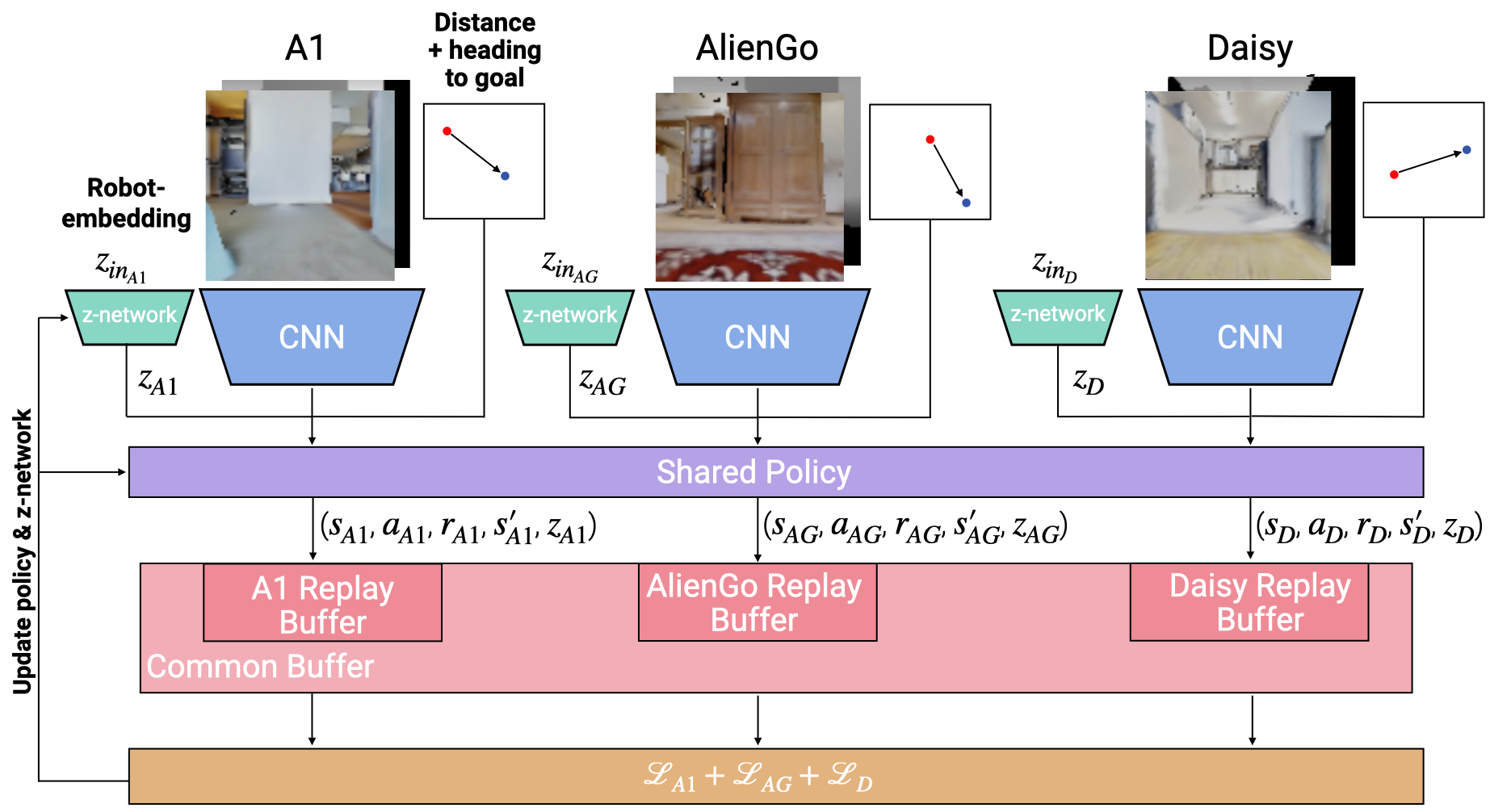}
    \caption{\small We train a high-level navigation policy for A1, AlienGo, and Daisy, along with robot-specific embeddings. The robots receive an egocentric RGB or Depth observation, the goal relative to the robot's current position, and a robot-specific embedding. Each robot contributes to a replay buffer, used for learning robot embeddings, and all robot buffers contribute to the update of the shared policy.}
    \label{fig:architecture}
      \vspace{-0.7cm}
\end{figure}

\xhdr{Learning high-level navigation policy:}
The z-network parameters $\psi$ are learned alongside the critic parameters $\phi$, by minimizing $\mathcal{L}_{critic}$, a variant of the update from \cite{haarnoja2018soft}:
\begin{align}
\vspace{-0.1cm}
    \hspace{-2pt} \mathcal{L}_{critic} = \mathds{E}_{(s,a,s',r,z\sim g_\psi)} \Big[Q_\phi (s, a, z) - \big(r + \bar{V}(s', z)\big) \Big]^2 \\ 
    \psi^{*}_{1}, \ldots, \psi^{*}_{N}, \phi^{*} = \argmin_{\psi_{1}, \ldots, \psi_{N}, \phi} \,\, \mathcal{L}_{critic}( \mathcal{D})
\end{align}
The partial derivative of $\mathcal{L}_{critic}$ with respect to $\phi$ uses samples from all robots, while when updating $\psi_i$ the partial derivative w.r.t $\mathcal{D}_{j \neq i}$ is 0. 
\vspace{-0.1cm}
\begin{align*}
    \frac{\partial \mathcal{L}_{critic}}{\partial \phi} = \frac{\partial }{\partial \phi} \mathds{E}_{\mathcal{D}} \Big[Q_{\phi}(s, a, z) - \big(r + \gamma \bar{V}(s', z)\big)\Big]^2 \\
    \frac{\partial \mathcal{L}_{critic}}{\partial \psi_i} = \frac{\partial }{\partial \psi_i} \mathds{E}_{\mathcal{D}_{i}} \Big[Q_\phi (s, a, z_i \sim g_{\psi_{i}}) - \big(r + \gamma\bar{V}(s', z_i)\big)\Big]^2
    \label{eq:update_z}
\end{align*}
This results in an update law that pools data across multiple robots to update the critic parameters $\phi$, but individual robot datasets $\mathcal{D}_i$ for updating robot-specific embedding $z_i$. The actor $\pi_\theta$ is updated using the common dataset $\mathcal{D}$, without updating robot embedding $\psi_i$, by minimizing KL divergence between $\pi_\theta$ and the predicted optimal Q-value, as in \cite{haarnoja2018soft}. Our training pipeline is summarized in Figure \ref{fig:architecture}.

In practice, we leverage Pytorch's \texttt{multiprocessing} to create parallel processes for multiple robots to collect experience in their own individual environments. 
By collecting data across multiple robots and environments for training both $\pi_\theta$ and $Q_\phi$, our training is more sample-efficient than training per robot, as well as robust to differences in robot dynamics and environments. 


Learned embeddings have been explored in literature before, for example in \cite{duan2016rl, li2020planning, peng2020learning, rakelly2019efficient}. However, none of these works generalize to multiple robot morphologies or use a high-dimensional image input as state. \cite{duan2016rl, peng2020learning, rakelly2019efficient} use dynamics randomization to create perturbed environments;  \cite{peng2020learning} input the dynamics parameters to learn the embedding, while \cite{duan2016rl, rakelly2019efficient} use robot state trajectories of past time steps as input. In contrast, we present a fully-learned embedding which does not need hand-defined dynamics features to learn an embedding for significantly different robot morphologies. We also compare our approach against \cite{duan2016rl} and \cite{peng2020learning} in Section \ref{sec:exp}, and show that we outperform both.

Learning a shared high-level policy allows robots to share data, significantly improving the sample-efficiency of learning. Our hierarchical controller makes training of the high-level policy even more expensive as for every action commanded by the high-level policy, the low-level policy takes multiple simulation steps to achieve the desired command. Hence, the number of the data points collected per high-level action is much smaller than the actual simulation steps. 
By pooling data across $N$ robots, we effectively get $N$ times as much data as we would if we collected data per robot. Our experiments show that our hierarchical structure with off-policy learning can learn navigation policies for multiple legged robots, and generalize to unseen robots and hardware.

\xhdr{Generalization to unseen robots.} 
Once learned, our navigation policies can be adapted to new robots by searching in the space of the learned 1-dimensional embedding $z_i$ which is input to the policy $\pi_\theta$. We aim to find a robot embedding $z^{*}_{test}$ that maximizes the long-term reward over trajectories $\tau$ of length $T$ induced on the test platform, using the learned policy $\pi_\theta(\cdot, z^{*}_{test})$: $z^{*}_{test} = \argmax_{z_{test}} \,\, \mathds{E}_{\tau} [\sum^{T}_{t=1} r_t]$.
We conduct a grid-search over $z_{test}$ in the interval $[-1, 1]$ to find the globally-optimal $z^{*}_{test}$ (within discretization error) in the space of learned embeddings. Approximate techniques like \cite{peng2020learning, yu2020learning} can be used for higher dimensional $z$. 


For successful navigation on new robots, the learned embedding should capture different properties that the robots demonstrate. For example, some robots might be slower at turning while others might be faster.
%
\begin{wrapfigure}{h}{0.2\textwidth}
\vspace{-0.4 cm}
    \centering
    \includegraphics[width=0.2\textwidth]{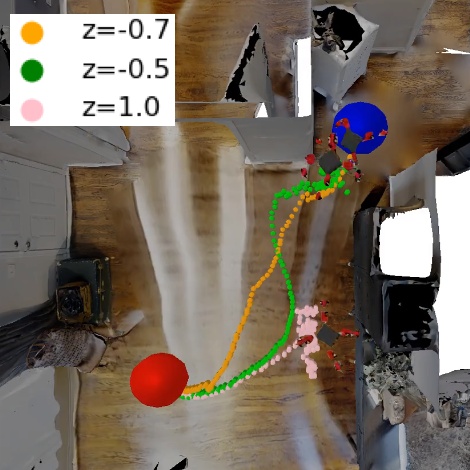}
    \caption{\small Different values of $z$ result in different CoM trajectories.}
    \label{fig:robots_z_latent}
    \vspace{-0.35 cm}
\end{wrapfigure}
We expect different values of $z$ to result in different motions of robots around obstacles, etc. 
Fig. \ref{fig:robots_z_latent} illustrates CoM trajectories in an environment with obstacles for different values of $z$. For $z=-0.7$, we see that 4-legged Daisy takes a relatively direct path to the goal, while for other $z$ values, the robot gets stuck on obstacles.

\subsection{Reward function}
\label{sec:reward}
Our reward function incentivizes path efficiency and safety of the robot:
\vspace{-0.2cm}
\begin{align}
    r_t(a_t,s_t) = R_{geo} + R_{coll} + r_{fall} + r_{success}
    \\
    R_{geo} = \Delta_{geo\_dist} \cdot k_{geo\_dist}, \, R_{coll} = -1.0 \text{ if collision}  \\
    r_{fall} = -5.0 \text{ if fall}, \quad  r_{success} = 10.0 \text{ if success}
\end{align}
$R_{geo}$ is the change in geodesic distance to the goal (shortest obstacle-free path to goal) from the previous step to the current step, scaled by a geodesic reward weight $k_{geo\_dist}$, a hyperparameter to encourage efficient paths. $R_{coll}$ penalizes the robot if there was a collision with the environment. $r_{fall}$ is the penalty for falling, applied if the robot falls in the step, at which point the episode is terminated. $r_{success}$ is the terminal reward given to the robot, if it reaches the goal.

\subsection{Training details}
\label{sec:training}
We use SAC+AE, a robust and sample-efficient method for learning from high-dimensional images for training our navigation policy and robot embeddings. The actor, which serves as the universal policy, consists of a 3-layer multi-layer perceptron (MLP) with a 1024-dimensional hidden layer, and outputs a vector of means and a covariance matrix that parameterize a Gaussian action distribution. The Q-function is represented by a 3-layer MLP with 1024 units. 

A convolutional autoencoder is used to provide the policy with visual features extracted from the input depth image. The encoder consists of 4 convolutional layers that map the input depth image to a 50-dimensional latent representation. The decoder consists of a fully connected layer followed by 4 deconvolutional layers that reconstructs the latent representation back to the original image. This reconstruction loss is used as an auxiliary objective during training; it improves sample efficiency and aids in training stability \cite{yarats2019improving}. 


%% file: experiments.tex
\section{Experimental evaluation}
\label{sec:exp}
We evaluate our proposed approach on two experimental setups: a set of cart-pole variants, and a set of legged robots. In both cases, we aim to learn generalizable policies on certain morphologies and test on new morphologies that were not seen during training. We first choose cart-pole environments as a simplified, but illustrative example of our framework, and compare with other approaches from literature. 
Our results show that our approach is faster at learning on a set of cart-poles, and better at generalization to unseen cart-poles than other approaches, like \cite{duan2016rl, peng2020learning}. 

Next, we tackle the much more challenging task of learning navigation policies on legged robots, and generalization to unseen robots. Our experiments on legged robots show that our approach can learn navigation policies across multiple legged robots, and generalize to hardware and unseen robots sample-efficiently. It outperforms baselines from literature like \cite{peng2020learning}, \cite{duan2016rl}, ablations of our approach, and an oracle point-agent policy (without dynamics) that uses $\text{A}^*$ on a map. 
\subsection{Experimental setup}
\label{sec:exp_setup}
 We compare our approach (\textbf{Learned-z}) to the following:

\noindent -- $\text{RL}^2$ with DDPG (\textbf{Meta-RL}): We use $\text{RL}^2$~\cite{duan2016rl} as a baseline that learns robot-specific contexts during training. At test time, context is inferred from state trajectories on the test robot. This baseline compares our learned robot-specific embedding to meta-learned embeddings from literature. 

\noindent -- Informed embedding (\textbf{Informed-z}): \cite{peng2020learning} learns a robot-specific embedding by providing the z-network with the robot's dynamics parameters. At test time, $z$ is optimized for the new robot, similar to our approach. This experiment compares our learned embedding, which does not require any prior knowledge of the dynamics parameters, to an informed embedding that has access to more privileged information. 

\noindent -- Semi-Informed embedding (\textbf{Semi-Informed-z}): In this experiment, we only give a sub-set of the varied dynamics parameters as inputs to the z-network during training (mass and offset of the cart-poles). Deciding the `right' input parameters can be challenging, especially when working with different robot morphologies. This experiment shows the sensitivity of \cite{peng2020learning} to an incorrect set of parameters. 
    
\noindent -- Fixed robot embedding (\textbf{Fixed-z}): In this ablation of our approach, we fix $z$ during training, giving each robot a distinct value. At test time, $z$ is then optimized for the new robot, similar to our approach. This experiment demonstrates that a learned embedding captures the landscape of the robot's dynamics better than fixed initialized embeddings. 

\noindent -- No robot-specific embedding (\textbf{No-z}): In this ablation of our approach, we learn a shared universal policy across different environments, with no robot-specific embeddings. At test time, the whole policy is fine-tuned on the new robot. This experiment demonstrates the need for robot-specific embeddings for sample-efficient generalization.

\subsection{Cart-pole experiments}
\begin{figure}[h]
\vspace{-0.3cm}
  \centering%
  \resizebox{\columnwidth}{!}{
  \renewcommand{\tableTitle}[1]{\large{#1}}%
  \setlength{\figwidth}{0.3\columnwidth}%
  \setlength{\tabcolsep}{1.5pt}%
  \renewcommand{\arraystretch}{0.8}%
  \renewcommand\cellset{\renewcommand\arraystretch{0.8}%
  \setlength\extrarowheight{0pt}}%
  
  
  \hspace{-0.25cm}\begin{tabular}{c c c c c}
  \multicolumn{3}{c}{\bfseries Train} & \multicolumn{2}{c}{\bfseries Test}\\
  \cmidrule(l{4pt}r{4pt}){1-3} \cmidrule(l{4pt}r{4pt}){4-5} \\
  CP1 & CP2 & CP3 & CP4 & CP5 \\ 
   \makecell{\includegraphics[width=\figwidth]{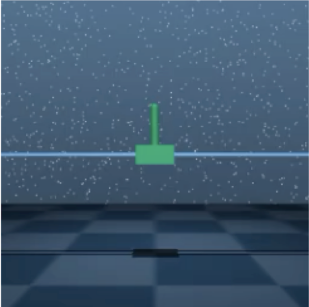}} &%
   \makecell{\includegraphics[width=\figwidth]{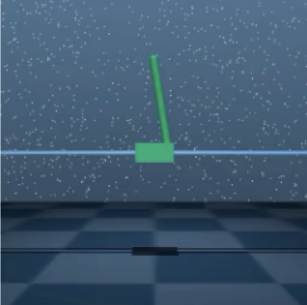}} &%
   \makecell{\includegraphics[width=\figwidth]{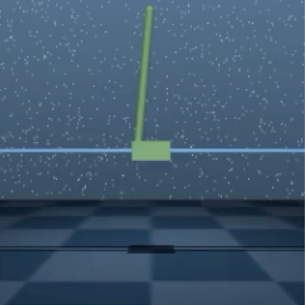}} &%
   \makecell{\includegraphics[width=\figwidth]{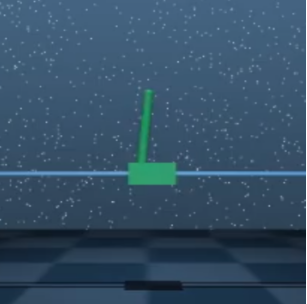}} &%
   \makecell{\includegraphics[width=\figwidth]{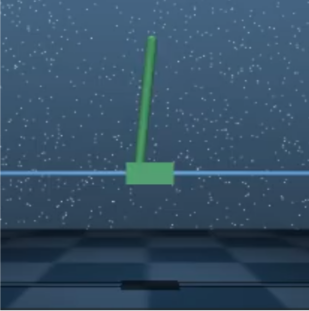}} \\
   \makecell{\includegraphics[width=1.3\figwidth]{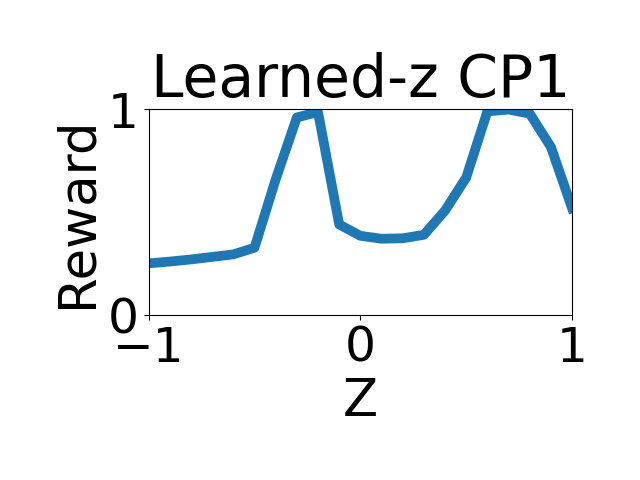}} &%
   \makecell{\includegraphics[width=1.3\figwidth]{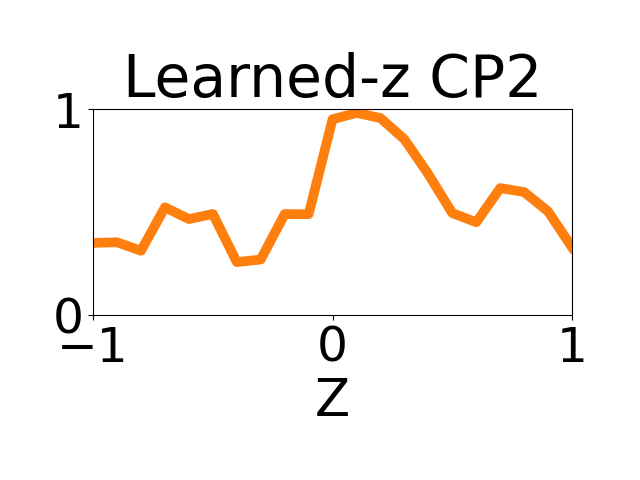}} &%
   \makecell{\includegraphics[width=1.3\figwidth]{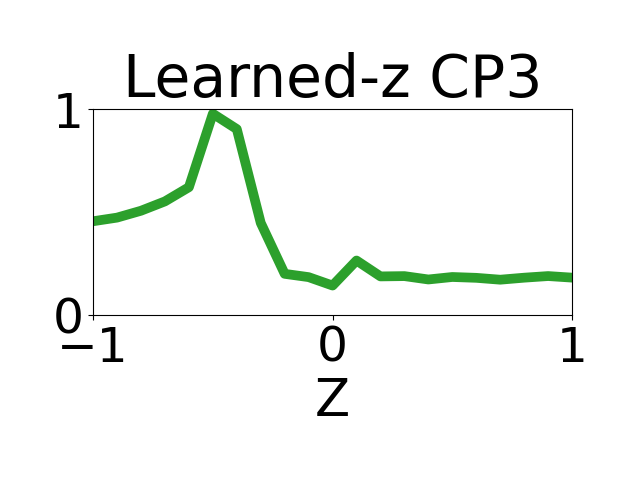}} &%
   \makecell{\includegraphics[width=1.3\figwidth]{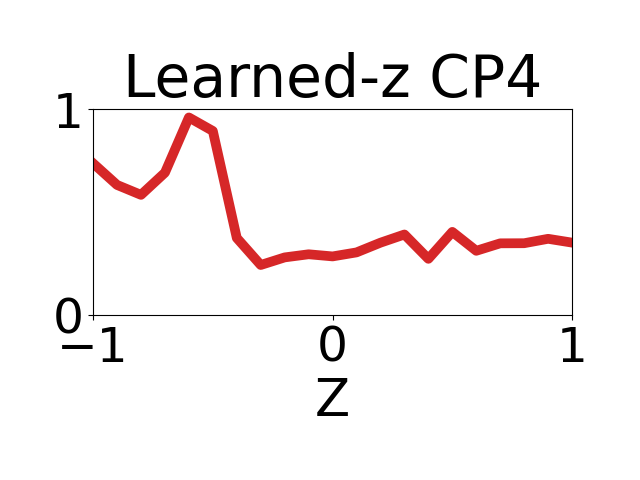}} &%
   \makecell{\includegraphics[width=1.3\figwidth]{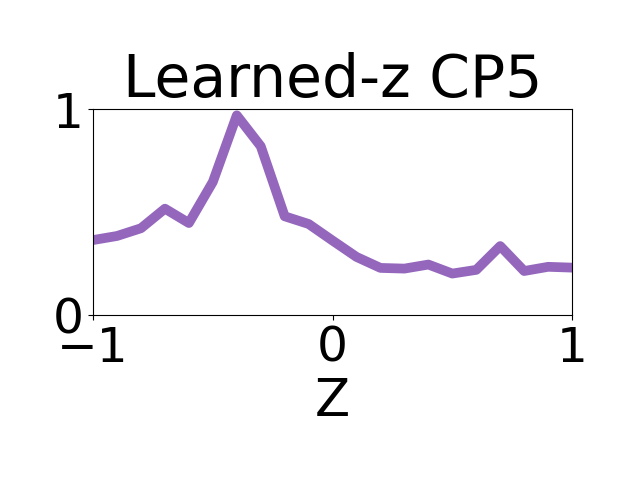}} \\
  \end{tabular}}
  \caption{\small (Top) Cart-pole robots used for training and testing. (Bottom) Performance of cart-poles for different $z$ in the learned latent space.}
  \label{fig:cp}
  \vspace{-0.19cm}
\end{figure}

We create a family of cart-poles in Mujoco \cite{Todorov2012MuJoCoAP} by changing the mass of cart, length of pole, and offset of pole from center (Fig. \ref{fig:cp}, Table \ref{tab:cp_params}). Test robots consist of one cart-pole that lies `between' the training cart-poles (CP5) and one that lies outside of the training distribution (CP4). Our aim is to test generalization of all approaches to both in and out-of-distribution robots. 

We visualize the learned latent space by sweeping over different values of $z$ and measuring performance (Fig. \ref{fig:cp}, bottom). All train cart-poles have a singular maximum, near the learned $z$, except CP1. Visualizing the latent space also shows that CP4 is close to CP3 in learned latent space, but out of distribution from train robots, while CP5 is between CP2 and CP3, as expected. This demonstrates that the learned embedding has captured dynamics properties of the robots.

\begin{table}[h]
\vspace{-0.5 cm}
\centering
\scriptsize
\caption{\small 
We create 5 variants of cart-poles with dynamics parameters shown below. We train on CP1-3, and test on CP4-5.}
\resizebox{0.7\columnwidth}{!}{
\label{tab:cp_params}
    \begin{tabular}{cccccc}
        \toprule
        \textbf{Parameter}  & \textbf{CP1} & \textbf{CP2} & \textbf{CP3} & \textbf{CP4} & \textbf{CP5} \\
        \cmidrule(l{0pt}r{0pt}){1-6}
        Mass & 0.1 & 1 & 2 & 1.5 & 1.5\\
        Pole Length & 0.5 & 1 & 1.5 & 0.75 & 1.25\\
        Pole Offset  & 0 & 0.15 & -0.15 & -0.1 & -0.1\\
        \bottomrule
    \end{tabular}
    }
    \vspace{-0.3 cm}
\end{table}

\begin{figure}[t]
  \centering%
  \resizebox{\columnwidth}{!}{
  \renewcommand{\tableTitle}[1]{\large{#1}}%
  \setlength{\figwidth}{0.3\columnwidth}%
  \setlength{\tabcolsep}{1.5pt}%
  \renewcommand{\arraystretch}{0.8}%
  \renewcommand{\cellset}{\renewcommand\arraystretch{0.8}%
  \setlength\extrarowheight{0pt}}%
  
  
  \hspace{-0.25cm}\begin{tabular}{c c}
   \makecell{\includegraphics[width=\figwidth]{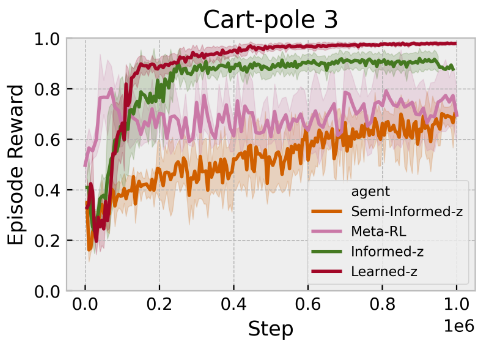}} &
   \makecell{\includegraphics[width=\figwidth]{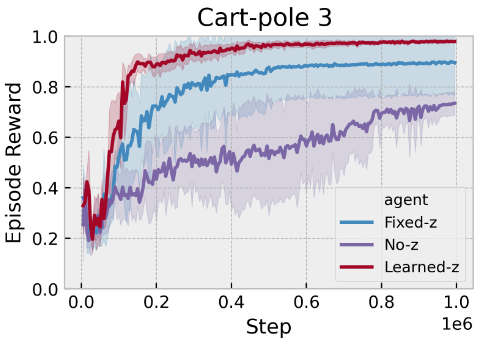}} \\
  \end{tabular}}
  \caption{\small Learning speed of different approaches, visualized on CP3. Our approach learns faster, and achieves a higher performance than all the other approaches. (Left) Comparison against context-aware approaches; (Right) Ablations of our approach.}
  \vspace{-0.5cm}
  \label{fig:cp_learning}
  \end{figure}

\begin{table}
\vspace{-0.5cm}
\centering
\caption{\small Cart-pole Evaluation (Episode reward)}
\label{tab:cp_eval}
\resizebox{\columnwidth}{!}{
    \begin{tabular}{cccccc}
        \toprule
        & \multicolumn{3}{c}{\bf{Train}} &  \multicolumn{2}{c}{\bf{Test}}\\ 
        \cmidrule(l{4pt}r{4pt}){2-4} \cmidrule(l{4pt}r{4pt}){5-6}
        \textbf{Experiment}  & \textbf{CP1 } & \textbf{CP2 } & \textbf{CP3 } & \textbf{CP4 } & \textbf{CP5 } \\
        \cmidrule(l{0pt}r{0pt}){1-6}
        Semi-Informed-z & 0.97\tiny{$\pm$0.02} & 0.74\tiny{$\pm$0.03} & 0.71\tiny{$\pm$0.04} & 0.42\tiny{$\pm$0.04} & 0.75\tiny{$\pm$0.01}\\
        Meta-RL & 0.99\tiny{$\pm$0.01} & 0.93\tiny{$\pm$0.02} & 0.70\tiny{$\pm$0.09} & 0.66\tiny{$\pm$0.11} & 0.83\tiny{$\pm$0.06}\\\
        Informed-z & 0.99\tiny{$\pm$0.01} & 0.90\tiny{$\pm$0.04} & 0.93\tiny{$\pm$0.02} & 0.79\tiny{$\pm$0.16} & 0.94\tiny{$\pm$0.04}\\
        No-z  & 0.99\tiny{$\pm$0.01} & 0.76\tiny{$\pm$0.08} & 0.80\tiny{$\pm$0.05} & 0.68\tiny{$\pm$0.10} & 0.85\tiny{$\pm$0.06}\\
        Fixed-z & 0.99\tiny{$\pm$0.02} & 0.86\tiny{$\pm$0.13} & 0.87\tiny{$\pm$0.19} & 0.77\tiny{$\pm$0.21} & \textbf{0.97\tiny{$\pm$0.03}}\\ 
        Learned-z & \textbf{1.00\tiny{$\pm$0.00}} & \textbf{0.98\tiny{$\pm$0.00}} & \textbf{0.98\tiny{$\pm$0.01}} & \textbf{0.94\tiny{$\pm$0.03}} & \textbf{0.97\tiny{$\pm$0.03}}\\
        \bottomrule
    \end{tabular}
    }
\end{table}
The training curves of our cart-pole experiments on CP3 are shown in Fig. \ref{fig:cp_learning}, and Table \ref{tab:cp_eval} condenses results for all cart-poles. 
\textbf{Meta-RL} does not learn to control CP3, and performs poorly on test cart-poles CP4 and CP5 ($0.66$ reward on CP4 and $0.83$ reward on CP5), showing that \textbf{Meta-RL} is unable to generalize to new unseen robots, with poorer performance on CP4 which is out of the training distribution. 
%
%

\textbf{Informed-z} is given the dynamics parameters of the robots (Table \ref{tab:cp_params}) during training, yet we notice that
\textbf{Learned-z} generalizes to test robot CP4 better ($0.94$ using \textbf{Learned-z} vs. $0.79$ using \textbf{Informed-z}). This indicates that for \textbf{Informed-z}, the information of dynamics parameters deteriorates generalization to out-of-distribution robots. \textbf{Semi-Informed-z} performs poorly at training and test robots, highlighting the sensitivity of \cite{peng2020learning} to the `right' set of dynamics parameters.

Among the ablation experiments of our approach, we observe that \textbf{Learned-z} generalizes better to the two test robots than \textbf{Fixed-z} and \textbf{No-z}. 
While both \textbf{Learned-z} and \textbf{Fixed-z} learn to control the training cart-poles, generalization to CP4 is improved when using learned robot embeddings ($0.94$ with \textbf{Learned-z} vs $0.77$ with \textbf{Fixed-z} and $0.68$ with \textbf{No-z}). 
\textbf{Informed-z} and \textbf{Fixed-z} are close to \textbf{Learned-z} in generalization to CP5, but \textbf{Learned-z} performs best when testing on out-of-distribution robot (CP4). 


    

\subsection{Simulation legged robot experiments}
\begin{figure}
    \centering
    \includegraphics[width=0.45\textwidth]{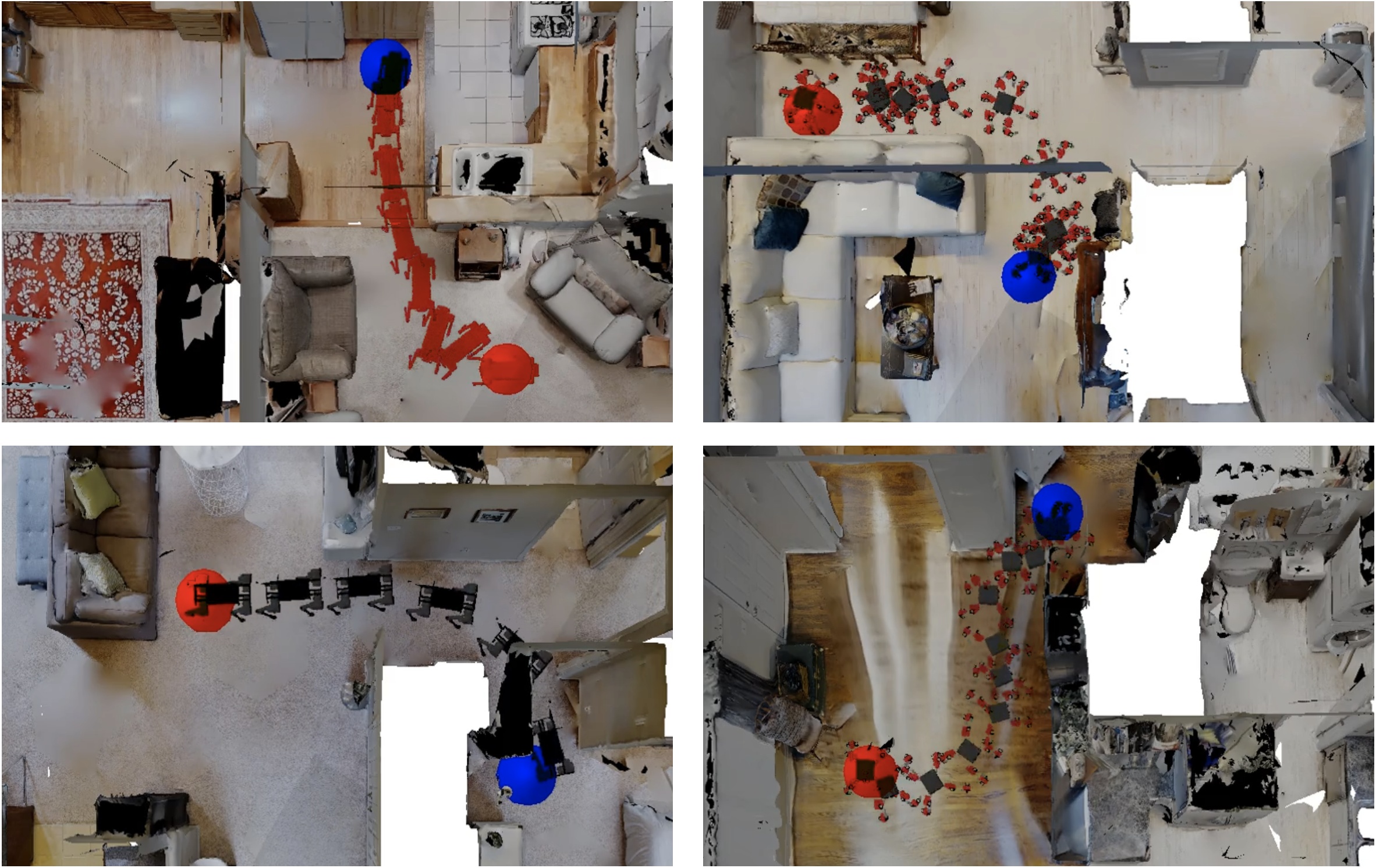}
    \caption{ \small Our approach learns policies capable of successfully navigating on different train (top) and test (bottom) robots.
    }
    \label{fig:robots_trajs}
    \vspace{-0.4cm}
\end{figure}

Next, we present experiments on legged robots in simulation (Fig. \ref{fig:robots_trajs}). In this setting, we use the hierarchical control structure described in Section \ref{sec:low_level} and learn a high-level navigation policy. Robots are trained from scratch in the iGibson \cite{xia2020interactive} simulator for a cumulative 1 million simulation steps across all robots. For all approaches, we train on 3 train robots (A1, Aliengo, Daisy), and evaluate generalization to 2 new robots (Laikago, Daisy-4-legged) in simulation. To compare the different approaches, we report the mean and standard deviation of SPL \cite{anderson2018evaluation} over 5 random seeds when the high-level policy is applied to both train and test robots. 

\begin{table}[h]\
\vspace{-0.35cm}
\centering
\caption{\small Legged robots' evaluation performance. Learned-z achieves the highest SPL when tested on previously unseen robots.}
\label{tab:robots_eval}
\resizebox{\columnwidth}{!}{
    \begin{tabular}{c c c c c c}
        \toprule
        & \multicolumn{3}{c}{\bf{Train}} &  \multicolumn{2}{c}{\bf{Test}}\\ 
        \cmidrule(l{4pt}r{4pt}){2-4} \cmidrule(l{4pt}r{4pt}){5-6}
        \bf{Experiment}  & \bf{A1} & \bf{AlienGo} & \bf{Daisy} & \bf{Laikago} & \bf{Daisy4}\\
        \cmidrule(l{0pt}r{0pt}){1-6}
        A* policy & 0.49\tiny{$\pm$0.07} & 0.45\tiny{$\pm$0.15} & 0.22\tiny{$\pm$0.10} & 0.38\tiny{$\pm$0.15} & 0.18\tiny{$\pm$0.06}\\
        Meta-RL & 0.08\tiny{$\pm$0.06} & 0.14\tiny{$\pm$0.12} & 0.03\tiny{$\pm$0.06} & 0.02\tiny{$\pm$0.03} & 0.07\tiny{$\pm$0.06} \\
        Informed-z & 0.57\tiny{$\pm$0.06} & 0.52\tiny{$\pm$0.05} & 0.43\tiny{$\pm$0.09} & 0.37\tiny{$\pm$0.11} & 0.40\tiny{$\pm$0.06} \\
        No-z & 0.48\tiny{$\pm$0.12} & 0.42\tiny{$\pm$0.05} & 0.33\tiny{$\pm$0.07} & 0.33\tiny{$\pm$0.13} & 0.22\tiny{$\pm$0.08} \\
        Fixed-z & \textbf{0.59\tiny{$\pm$0.07}} & \textbf{0.59\tiny{$\pm$0.14}} & 0.40\tiny{$\pm$0.10} & 0.47\tiny{$\pm$0.09} & 0.38\tiny{$\pm$0.11} \\
        Learned-z & \textbf{0.59\tiny{$\pm$0.07}} & 0.54\tiny{$\pm$0.04} & \textbf{0.48\tiny{$\pm$0.14}} & \textbf{0.48\tiny{$\pm$0.08}} & \textbf{0.49\tiny{$\pm$0.08}} \\
        \bottomrule
        \vspace{-0.5cm}
    \end{tabular}
    }
\end{table}

\xhdr{Zero-shot transfer of $\text{A}^*$ policies without dynamics information:}
We create an oracle point-based policy which uses a map of the environment to find a near-optimal collision-free path to goal. 
The policy samples a path to goal using $\text{A}^*$, with near perfect performance ($0.92$ SPL) on an idealized cylinder agent. However, when applied on the legged robots, the performance drops significantly (Table \ref{tab:robots_eval}),  with the most significant drop seen on 4-legged Daisy ($0.18$ SPL). This is because the oracle policy does not account for robot dynamics, and the robots often fall, get stuck around an obstacle, or exceed maximum allowed steps before reaching the goal. This emphasizes that knowledge of the low-level dynamics of legged robots is crucial for effective navigation.


\xhdr{Context-aware comparisons:} Table \ref{tab:robots_eval} shows that relative to other context-aware approaches, our approach is able to achieve a higher SPL on the test robots while retaining high SPL on the train robots. \textbf{Meta-RL} achieves poor performance on all robots ($<$$0.15$ SPL), with the robots falling over in most episodes. This result highlights the difficulty in generalizing \textbf{Meta-RL} to more complex dynamical systems such as legged robots. Although \textbf{Informed-z} is given the dynamics parameters of the robots (Table \ref{tab:robot_params}) during training, it does not generalize to Laikago and 4-legged Daisy as well as \textbf{Learned-z} ($0.37$ SPL vs. $0.48$ SPL for Laikago, $0.40$ SPL vs. $0.49$ SPL for 4-legged Daisy). This illustrates the difficulty in selecting the `right' set of dynamics parameters across robots of different morphologies.

\textbf{Learned-z} and \textbf{Fixed-z} slightly outperform \textbf{No-z}. However, on generalization robots we see an improvement when using \textbf{Learned-z} especially on 4-legged Daisy ($0.49$ SPL with \textbf{Learned-z} vs $0.38$ SPL with  \textbf{Fixed-z} and $0.22$ SPL with  \textbf{No-z}). Since 4-legged Daisy is the most `different' robot from the train robots, we again observe that our approach performs best at out-of-distribution generalization. 

\begin{table}[ht]
\vspace{-0.2 cm}
\centering
\scriptsize
\caption{\small 
Dynamics parameters of the robots used as input to the z-network for \textbf{Informed-z}.}
\resizebox{\columnwidth}{!}{
\label{tab:robot_params}
    \begin{tabular}{cccccc}
        \toprule
        \textbf{Parameter}  & \textbf{A1} & \textbf{AlienGo} & \textbf{Daisy} & \textbf{Laikago} & \textbf{Daisy4} \\
        \cmidrule(l{0pt}r{0pt}){1-6}
        Mass (kg) & 12.46 & 20.64 & 24.76 & 20.74 & 17.62\\
        Leg Length (m) & 0.4 & 0.5 & 0.45 & 0.5 & 0.54\\
        Number of Legs  & 4 & 4 & 6 & 4 & 4\\
        \bottomrule
    \end{tabular}
    }
    \vspace{-0.4 cm}
\end{table}

\subsection{Hardware experiments}
\label{sec:hardware}

Finally, we transfer our learned navigation policy from simulation to a real A1 quadrupedal robot. The robot is equipped with a RealSense depth and tracking camera for capturing depth images of the environment, as well as localizing itself with respect to its start location. The goal is defined with respect to the start point of the robot, and transformed to the current robot position at each robot step. 

On hardware, we do not optimize $z$, and instead use the learned $z$ from simulation for the A1 robot. The high-level policy operates at 4Hz, taking as input the depth camera image as seen by the robot, relative goal location, and learned $z$ from simulation. It commands a desired CoM velocity to the low-level controller, which operates at 250Hz. The stance controller for the A1 robot is a whole-body MPC controller, similar to \cite{peng2020learning}, and the swing legs use IK (Section \ref{sec:low_level}). 

\begin{figure}
    \centering
    \includegraphics[width=0.5\textwidth]{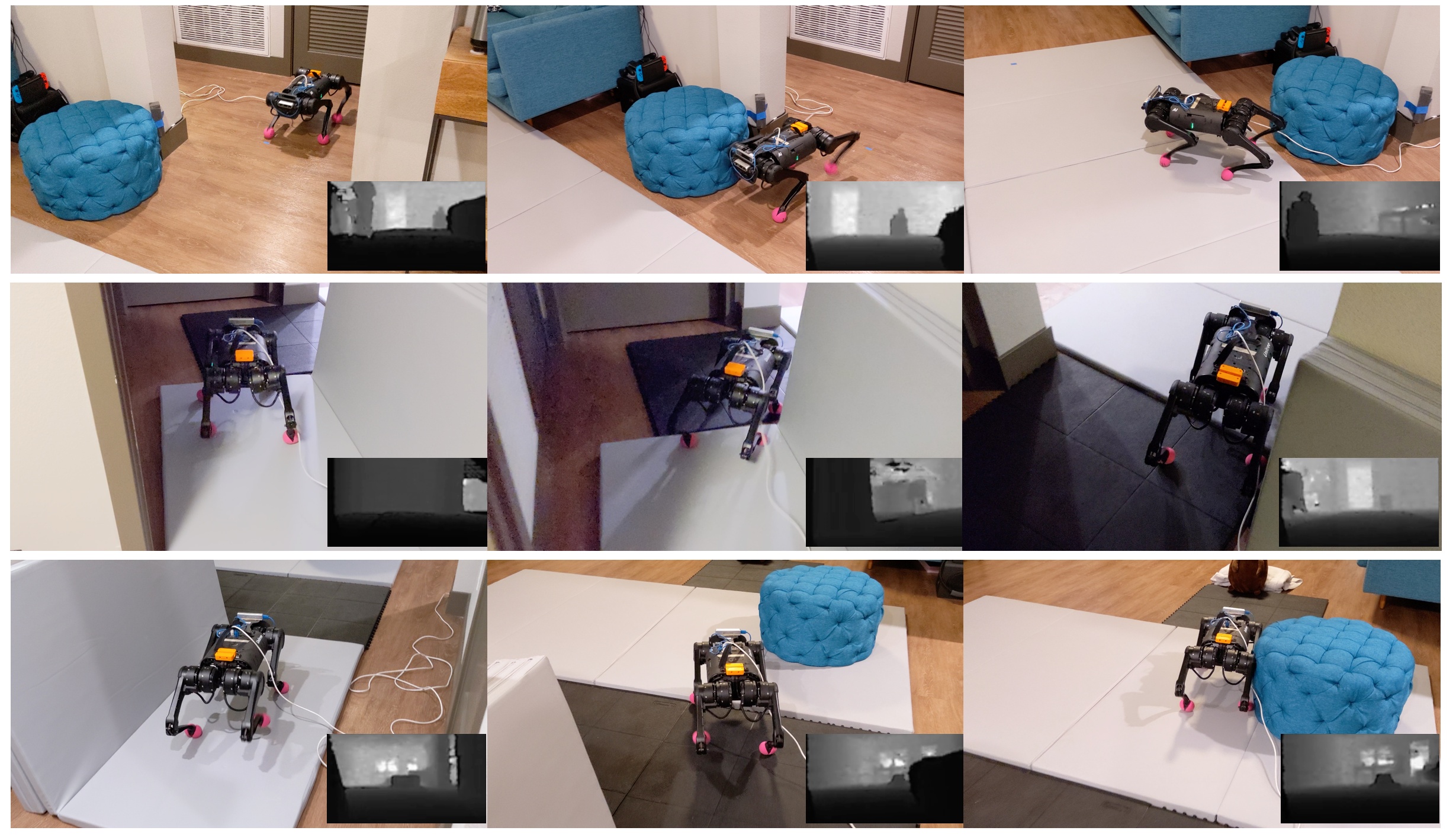}
    \caption{\small Snapshots of the A1 robot navigating a new home. The robot goes through hallways, and avoids obstacles to reach goals. Depth images, as seen by the robot, are shown in the bottom right corner of each frame.}
    \label{fig:hardware}
    \vspace{-0.65cm}
\end{figure}

Our hardware experiments show the A1 robot moving through hallways to reach the next room, and avoiding obstacles to reach a goal (Figure \ref{fig:hardware}). These experiments show that our proposed approach can learn robust high-level navigation policies that can transfer from simulation to real world, without any fine-tuning. During our experiments, we noticed that the A1 robot would turn very close to obstacles, and sometimes hit corners. This was caused by the hardware having different dynamics than the simulation, for example, the real robot turns slower than the simulation, and hence can hit obstacles even when the policy planned not to. This performance can be improved by fine-tuning the learned embedding on hardware, and searching for a $z$ that results in wider motions on the real robot. Sample-efficiently optimizing $z$ that can generalize to multiple goals on hardware is challenging, and we leave this to future work.

%% file: conclusion.tex
\section{Conclusion}
In this work, we present a framework for learning hierarchical navigation policies for legged robots. 
Our policy shares data across multiple robots during learning, and learns a high-level navigation policy with a shared universal navigation policy across robots, and robot-specific embeddings, used for specialization.
The learned embedding captures different dynamical properties, such as turning radius and walking speed of different robots, and shows promising results on hardware, and robots that were never seen during training. 
This work opens up future avenues for large-scale research on legged platforms for indoor navigation.

%% file: acknowledgements.tex
\section{Acknowledgements}
\scriptsize{
The Georgia Tech effort was supported in part by NSF, AFRL, DARPA, ONR YIPs, ARO PECASE. JT was supported by an NSF Graduate Research Fellowship under Grant No. DGE-1650044 and a Google Women Techmaker’s Fellowship. The views and conclusions are those of the authors and should not be interpreted as representing the U.S. Government, or any sponsor.

\noindent \textbf{License for dataset used} Gibson Database of Spaces. License at \url{http://svl.stanford.edu/gibson2/assets/GDS_agreement.pdf}
}